\documentclass{article}

\usepackage[preprint]{corl_2021} 

\usepackage{graphicx}
\usepackage[T1]{fontenc}

\title{Reimagining an autonomous vehicle}

\author{
  Jeffrey Hawke, Haibo E, Vijay Badrinarayanan, Alex Kendall\\
  Wayve Technologies Ltd., London, United Kingdom\\
}

\begin{document}
\maketitle


\begin{abstract}
    The self driving challenge in 2021 is this century's technological equivalent of the space race, and is now entering the second major decade of development. Solving the technology will create social change which parallels the invention of the automobile itself. Today's autonomous driving technology is laudable, though rooted in decisions made a decade ago. We argue that a rethink is required, reconsidering the autonomous vehicle (AV) problem in the light of the body of knowledge that has been gained since the DARPA challenges which seeded the industry. What does AV2.0 look like? We present an alternative vision: a recipe for driving with machine learning, and grand challenges for research in driving.
\end{abstract}



\section{Introduction}

The modern autonomous driving industry draws its roots from the DARPA Urban Challenge in 2007 \cite{10.5555/1822544}. Since then, we have seen rapid growth of research in both academic and industrial settings. The majority of this has been in industry, as many technical problems are beyond the resources of academic laboratories. The challenge has been harder than anticipated, with a widely held view that the industry has been humbled by this experience. Nevertheless, the technology continues to advance and it is likely that we are beyond Gartner's \emph{trough of disillusionment} \cite{hype}, supported by increasing public milestones and further investment.

A key question this provokes is simply: are autonomous vehicles (AVs) solved? All indications point to `no'. AV technology has not easily scaled thus far, hence there must be some gap. A challenge in assessing this question is that the industry is tight-lipped, due to the competition and the economic opportunity on offer. Our assessment in this paper is through our own experience developing autonomous driving systems, with additional insight from industry publicity, research, and conversations.

We assess the primary factors for this lack of delivery to be: 1) technical scalability, 2) safety-critical engineering effort, 3) unit economics (profitability of an AV based on utilisation and costs), 4) regulation. Of these, we argue that the primary factor is \emph{technical scalability}. By this, we mean the ability of the decision-making software systems to generalize to new situations quickly with sufficient performance for deployment. First, let us discuss the other factors.

Making a safety-critical decision-making system of any type is no small engineering feat, but it is one that may be solved with time and investment. Similarly, we do not consider regulation to be the sole barrier. Regulators worldwide have shown a willingness to support autonomous driving \cite{usa,germany,uk}, and we consider this simply a matter of time, pending technological performance. Finally, we do not believe that unit economics are a dominant factor. While retrofit AVs are costly \cite{cost}, these costs have reduced with maturity and volume such that it could be addressed if sufficient utilisation and scale was technically possible. This scale problem is that one that venture capital models are very familiar with: invest until scale is achieved, bring costs down with scale, then extract profit. Given the availability of capital, we can discount this as the primary factor.

This leaves us with technical scalability as the underlying problem, which suggests that we have yet to really solve autonomous driving. Why might this be? We propose that we have been solving the problem in the wrong direction, and there is ample opportunity for research to further the field.

The technology that brought us from the DARPA era to today can be described as solving specialized general intelligence by combining components of even narrower intelligence. We refer to this as AV1.0. This was necessary given the constraints and goals of 2007, but arguably limits generalization to complex situations due to brittle decision-making. The alternative lies in AV2.0: a complete rethink of what it means to architect a decision-making system for driving with machine learning.

\section{AV1.0: The limitations of a classical autonomous driver}
\label{sec:citations}

AVs today are designed around the same deliberative robotics architecture, which is arguably an expansion of the sense-plan-act paradigm as in \cite{1218700}. For a detailed assessment, we refer the reader to surveys such as \cite{9046805, BADUE2021113816}. We argue that the key gap is decision-making intelligence. To assess this, let's consider the problem addressed by classical AV components, and limitations on decision-making.

\begin{enumerate}
    \item \textbf{Sensing}. 
    \emph{Problem}: Can we observe sufficient information about the environment to make the correct driving decision? 
    \emph{Limitation}: Sensor development has focused on increasing range and fidelity, arguably due to the industry shift towards autonomous trucking business models on highways. Improved fidelity is certainly helpful, but this is not a limitation for non-highway applications as there is sufficient raw information present to make a decision.
    
    \item \textbf{Scene Representation}: building a hand-crafted representation of the world.
    \begin{enumerate}
        \item \textbf{Localization \& mapping}. 
        \emph{Problem}: Where is my robot in a known map? This simplifies real-time decisions by transferring part of the representation offline with clean, curated data \cite{8025618}. 
        \emph{Limitation}: 1) maintenance complexity as the world changes, and 2) there is an unclear need if downstream decision-making were sufficiently robust.
        
        \item \textbf{Perception}. 
        \emph{Problem}: Can we extract sufficient context from the raw data for decision making? 
        \emph{Limitation}: Modern perception algorithms are extremely good, and a supervised approach is largely sufficient given data, time, and resources. We can perceive almost anything desired. Rather, the limitation is the hand-crafted representation itself. Do we have the information necessary for the decision? Furthermore, reducing sensor data to symbolic data may not generalize well, for example, failing to correctly interpret a pedestrian walking a bicycle given a hand-crafted taxonomy.
        
        \item \textbf{Behavior prediction}. 
        \emph{Problem}: Many decisions require an estimate of future state, which is complex due to the dynamic scene. 
        \emph{Limitation}: Prediction is sensitive to upstream error and has a dependency with planning. This means that an isolated prediction system will always have some representational error.
    \end{enumerate}
    \item \textbf{Planning}: decision-making, given a world representation.
    \begin{enumerate}
        \item \textbf{Behavior planning} 
        \emph{Problem}: how should an autonomous vehicle achieve its goal (e.g., route) given the current representation. 
        \emph{Limitation}: 1) it is extremely difficult to determine whether the input representation is necessary and sufficient for all decisions; 2) it is difficult to separate this from behavioral prediction, as decisions made by the planner will influence other actors and thus the prediction; 3) behavioral planners can be described as a highly engineered \emph{expert system} \cite{10.5555/1671238}, which are well known for being brittle. Despite this immense human effort, we have reached the same conclusion as thirty years earlier: symbolic expert systems are inherently limiting \cite{brooks1990elephants}.
        
        \item \textbf{Motion planning}. 
        \emph{Problem}: generate a metric trajectory for a short horizon, given local constraints \cite{7490340}. 
        \emph{Limitation}:  This works well, but cannot overcome limitations of upstream decisions or hand-crafted constraints.
    \end{enumerate}
    
    \item \textbf{Control}. 
    \emph{Problem}: How does a vehicle execute some trajectory? 
    \emph{Limitation}: Vehicle dynamics and control are relatively well understood with remaining challenges in low friction.
\end{enumerate}

With the exception of behavior prediction and planning, we consider the majority of these to be sufficiently mature for driving based on the success of respective benchmarks. Incrementally increasing perception performance does not enable a human observer to now make an effective judgement of how a car should drive given this increase in information. While further gains may be had, we do not believe any of these areas will offer a step change to unlock scalable driving. 

What evidence exists for behavior prediction and planning as the limiting factor? We refer to the increasing focus of the AV industry on research associated with motion prediction. Since 2018, we have seen a flurry of published research and datasets. Examples include Nutonomy's NuScenes dataset \cite{nuscenes2019}, Waymo Open Motion Dataset \cite{sun2020scalability}, the Lyft Prediction Dataset \cite{lyft2020}, and Argo's Argoverse \cite{Argoverse}. A substantial focus of these is prediction, evidenced by \cite{pmlr-v100-chai20a, hu2021fiery, hendy2020fishing, zhao2020tnt}.

Given this landscape, there are two possible conclusions. Either, solving \emph{behavior prediction \& planning} as defined by these boundaries will enable self-driving, or we need a rethink of this decomposition to achieve an autonomous future. We think the former is unlikely to be sufficient alone.



\section{AV2.0: The solution of a data-driven learned driver}

\begin{figure}
    \centering
    \includegraphics[width=0.9\columnwidth]{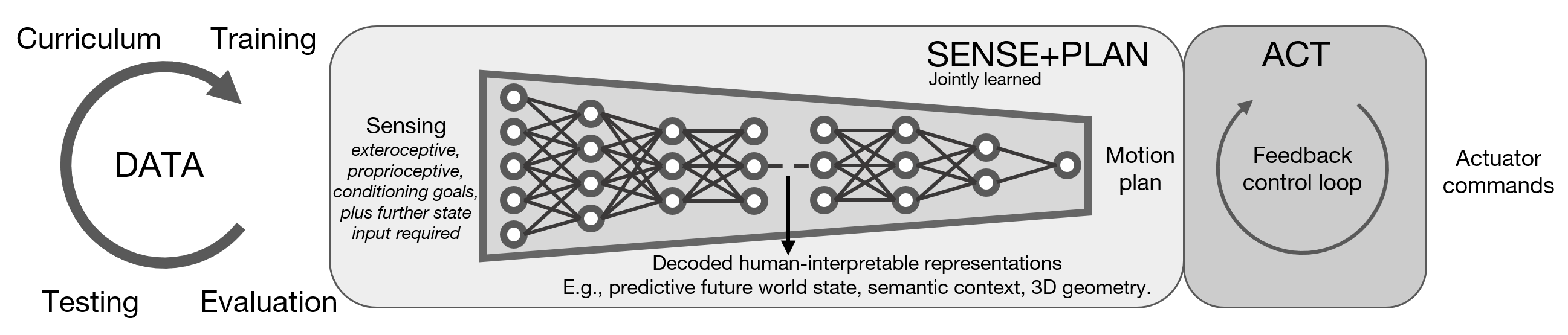}
    \caption{How should we reimagine an autonomous vehicle, given the progress the scientific and engineering academy has made since 2007? We propose that the classical deliberative architecture needs a rethink. We reflect on the long-held `sense-plan-act' paradigm (substantially simpler than the robot architectures used in modern autonomous vehicles), and pose the joint sensing and planning problem as one that may be solved by data.}
    \label{fig:reimagined-arch}
\end{figure}

\subsection{Solving driving with data}
With the challenges of the classical approach in mind, how should we best solve driving? It is hard to combine hand-crafted abstraction layers in a complex decision-making system without brittleness.

We have seen good progress in similar fields by posing a complex problem as one that is able to modeled end-to-end by data. Examples include natural language processing with GPT-3 \cite{NEURIPS2020_1457c0d6}, and in games with MuZero \cite{muzero} and AlphaStar \cite{alphastar}. In these problems, the solution to the task was sufficiently complex that hand-crafted abstraction layers and features were unable to adequately model the problem. Driving is similarly complex, hence we claim that it requires a similar solution.

The solution we pursue is a holistic learned driver, where the driving policy may be thought of as learning to estimate the motion the vehicle should conduct given some conditioning goals. This is different to simply applying increasing amounts of learning to the components of a classical architecture, where the hand-crafted interfaces limit the effectiveness of data \cite{unreasonable-data}.

At a high level, the route to a learned driver may be expressed very simply as the following:

\begin{enumerate}
    \item Frame the driving problem as one that may be solved by data.
    \item Build a source of data, for sampling and curation of data at sufficient scale and diversity.
    \item Build a data engine, which is able to train and iterate effectively on this shifting corpus.
    \item Build an experimental environment to explore the problem, in simulation and reality.
    \item Iteratively improve driving, through changes in modeling, data, and problem formulation.
\end{enumerate}


The key shift is \emph{reframing the driving problem as one that may be solved by data}. This means removing abstraction layers used in this architecture and bundling much of the classical architecture into a neural network, as outlined in Figure \ref{fig:reimagined-arch}. This is frequently depicted as the `end-to-end' approach to driving \cite{pomerleau1989alvinn}. However, this does not mean we believe the driving problem is simply solved by attaching a neural network directly to vehicle actuators. For instance, classical control methods together with learned representation and abstraction layers are still very effective at trajectory following.

\subsection{Grand challenges for learned driving}
With this move away from modularity, we arrive at a deep learning model that encapsulates much of the function of a classical robotics architecture.
We suggest posing driving as a model-based policy learning problem such as \cite{hafner2020mastering}: one where we learn both a predictive model of the world (conditioned on ego-vehicle actions) and a model-based policy. We may also retain many of the key inductive biases for driving \cite{casas2021mp3, Zhoueaaw6661, hawke2020urban} as part of this framework. However, this means more than learning a representation independently of control \cite{phillips2021deep}. To build this next-generation autonomy architecture, we suggest grand challenges for driving research.

Work on many of these challenges exists in current research across robotics and machine learning. However, many of these wider research threads have yet to focus on autonomous driving.

\begin{enumerate}
    \item \textbf{Vehicle adaptability}: deployment with different sensors, vehicle platforms, and use cases. A learned driver needs to be agnostic to sensor types, sensor rigs, and robot dynamics. It should adapt the collective experience of all driving to its vehicle, e.g., \cite{ghadirzadeh2021bayesian, kumar2021rma}. This enables using a policy or adaptation modules across heterogeneous fleets and cultures.
    
    \item \textbf{Modeling real-world complexity}: multi-agent, dynamic environments where future state is conditioned on current action \cite{7995949}. To model the true scene dynamics, a learned driver needs to model prediction and planning jointly, where prediction is conditioned on action \cite{hafner2020mastering}. This enables learning from imagined states which increases sample efficiency and policy robustness. 
    
    \item \textbf{Learning from accessible off-policy data}: train and evaluate from scalable data sources. Currently, much research is focused on online and on-policy learning, which is unlikely to have impact beyond toy reasoning problems. We need to learn in a way that is feasible \cite{levine2020offline}, as on-policy scaling is extremely slow and costly \cite{waymo}. A particular need is effective off-policy evaluation to compare driving decisions, e.g., \cite{chandak2021universal}.
    
    \item \textbf{Safety under uncertainty}: know when and what we don't know. In addition to common learning failures \cite{amodei2016concrete}, a learned driver needs to identify when possible decisions are uncertain \cite{jain_khetarpal_precup_2021, 7995949, mcallister2017concrete}. This enables robust performance and a safety case, by revoking control to simpler but robust systems to safely halt the vehicle when no clear decision exists.
    
    \item \textbf{Interpretability of failures}: disentangle the causal factors in decisions. A learned driver needs to be able to identify causal factors \cite{9363924} in decision-making where the encoding is sufficiently disentangled \cite{higgins2018towards}. E.g., failing to stop for a traffic light by incorrectly associating it to our vehicle is distinct from failures to see it due to weather. This interpretability enables development and will enable verification \cite{46160}.
    
    \item \textbf{Generalization to new situations}: a generalizable policy requires complete but lean generalizable representations. A learned driver needs to generalize to a new distribution every time it goes on the road \cite{pmlr-v119-zhang20t}, requiring repurposable driving knowledge \cite{pathakICMl17curiosity}. For example, our driver needs to reason about a multitude of different bicycle types: road bikes, cargo bikes, and commuter e-bicycles share many similarities (e.g., use of dedicated cycle lanes), but have large dissimilarities in visual appearance and speed. Additionally, a learning-based AV stack is transferable to a new geographic environment with relative ease, if the representation generalizes. This is a revolution: no classical AV stack can do this.
    
    
    \item \textbf{Driving reward}: optimization criteria for society's changing driving needs. A learned driver may start with supervision, but it may be difficult to go beyond human performance. Additionally, driving requirements are unlikely to be static, thus we must adapt to changing driving regulations as AVs become the dominant road user. For example, the following distance may be decreased to enable increased vehicle density on a largely automated road. We won't have the opportunity to regather expert supervisory data, hence we need a learning signal to shape driving behavior accordingly. This is largely unexplored \cite{lillicrap2016continuous, kendall2019learning}.
\end{enumerate}

In our opinion, truly solving learned autonomous driving requires solving each of these challenges. There are wider questions of using deep learning for decision-making which apply more broadly than driving, including ethics. We consider these beyond the scope of this paper.

\section{Learned driving is not uniquely challenged by safety-critical requirements}
\label{sec:safety}

Do safety-critical requirements mean that learned driving is impossible? This medium is insufficient for a comprehensive analysis, but in short, no. There is no inherent safety reason why we should not pursue a data-driven driver. This hinges on a number of key theses.

\begin{enumerate}
    \item Safety assurance of an AV depends on (1) the design of an architecture which includes, but is not limited to, the neural net, plus (2) an engineering effort in verification and validation. Core safety tools should apply with some thoughtful adaptation \cite{koopman2016challenges}.
    \item Within the broader architecture of the AV, redundant safety may be achieved with interpretable methods designed to identify and resolve specific failure modes. These methods cannot offer the generalized decision-making that a neural net can, but they are able to ensure that a vehicle will not cause harm in a very specific way \cite{Reason768}.
    \item Scalable safety arguments will use a large non-stationary corpus of empirical evidence. This is not unique to learned driving, and there is precedent in other domains such as in medicine \cite{drugs}. We anticipate that AVs may benefit from similar methods to quickly revalidate the learned driver as part of a wider system, however, this is an open question.
\end{enumerate}

\section{What's required to address these grand challenges?}
\label{sec:conclusion}

To see significant progress in the twenty-first century's space race, we suggest the following.

For the academic community, we encourage exploring the full \emph{embodied intelligence} problem space \cite{corke2020can} beyond just modeling, including data curriculum, sensor configuration, and robot geometry. Within modeling research, we encourage even more focus on off-policy learning and evaluation, to make better use of available data with increased research impact.

For the industry research community, we encourage sharing of data of rare events and data curricula for the full driving problem for benchmarks. Today most datasets describe sub-problems under the assumption that the set is sufficient for solving the whole \cite{nuscenes2019, sun2020scalability, lyft2020, Argoverse}, which we consider insufficient for driving research. We also encourage continued open collaboration with academia by sharing progress and the insights on the problems gathered from real world testing at scale.

Beyond this, we see a need for increased availability of holistic simulators. Tools such as CARLA \cite{osinski2020carla} are still nascent, and research benchmarking standards require more efforts to mature.

In solving the driving problem together, we have the potential to unlock both great societal value and also to discover and create embodied intelligence in the open world. 
We encourage research efforts on these grand challenges, to tackle one of the most complex problems of technical scalability.

\clearpage


\bibliography{main}  

\end{document}